\documentclass[]{gtech}
\usepackage{amssymb}
\usepackage{multirow}
\usepackage{bigdelim}
\usepackage{todonotes}
\usepackage{longtable}
\usepackage{tabularray}
\usepackage{wrapfig}
\usepackage[most]{tcolorbox} 
\usepackage{hyperref} 

\hypersetup{
  colorlinks=true,
  citecolor=green!50!black,
  linkcolor=red!60!black,
  urlcolor=green!50!black
}
\usepackage{xcolor}
\usepackage{url}
\usepackage{mathtools}
\usepackage{amsmath}
\usepackage{amsfonts}
\usepackage{booktabs}
\renewcommand{\title}[1]{\newcommand{\titlelist}{{\huge\fontfamily{optimistic}\selectfont #1}}}

\definecolor{mingtokblue}{HTML}{0369FF}
\newcommand{\model}{\texttt{\textbf{\textcolor{mingtokblue}{MingTok}}}}
\newcommand{\unifiedmodel}{\texttt{\textbf{\textcolor{mingtokblue}{Ming-UniVision}}}}

\definecolor{prompt}{HTML}{5f84e4}
\definecolor{img}{HTML}{820100}

\usepackage{arydshln}

\usepackage{colortbl}
\usepackage{amssymb}
\usepackage{pifont}
\usepackage{booktabs,multirow}
\usepackage{makecell}
\usepackage{tabulary}
\usepackage{fontawesome5}
\usepackage{bbding}
\usepackage{multicol}

\newlength\savewidth

\def\huggingface{\raisebox{-1.5pt}{\includegraphics[height=1.05em]{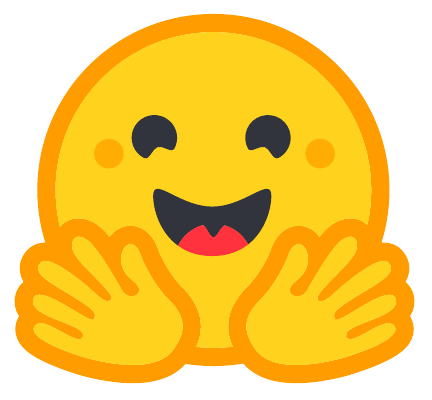}}}
\def\modelscope{\raisebox{-1.5pt}{\includegraphics[height=0.85em]{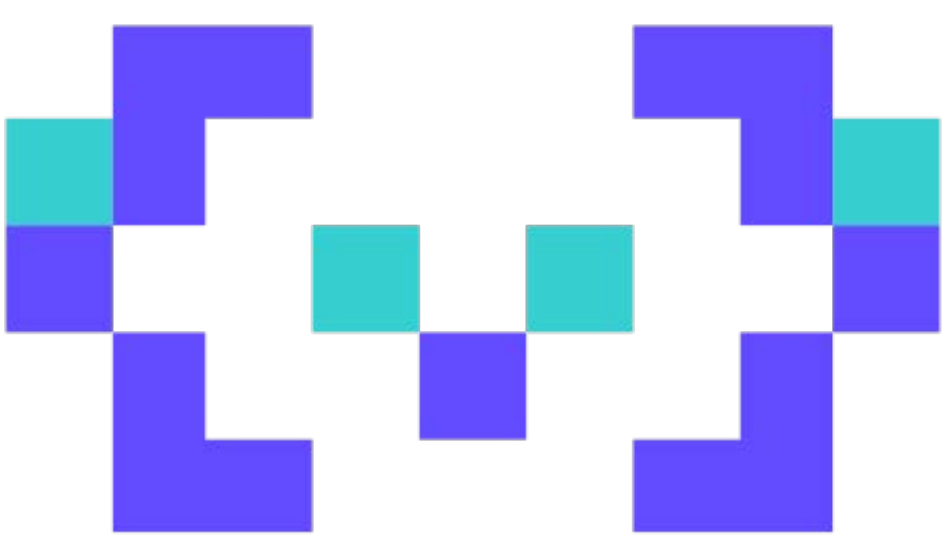}}}
\def\github{\raisebox{-1.5pt}{\includegraphics[height=1.05em]{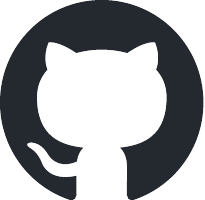}}}

\title{\textcolor{mingtokblue}{Ming-UniVision}:\\Joint Image Understanding and Generation with a Uni{f}{i}ed Continuous Tokenizer}

\author[*]{Inclusion AI, Ant Group}
\contribution[*]{See Contributions section (Sec.~\ref{sec:contri}) for full author list.}

\abstract{\fontsize{11pt}{12pt} 
\textit{Visual tokenization remains a core challenge in unifying visual understanding and generation within the autoregressive paradigm. 
Existing methods typically employ tokenizers in discrete latent spaces to align with the tokens from large language models, where the quantization errors can limit semantic expressiveness and degrade the capability of vision-language understanding. 
To address this, we introduce \model{}, a new family of visual tokenizers with a continuous latent space, for unified autoregressive generation and understanding.
While understanding tasks favor discriminative high-dimensional features, generation tasks prefer compact low-level codes. 
Thus, to reconcile these competing demands, \model{} adopts a three-stage sequential architecture involving low-level encoding, semantic expansion, and visual reconstruction.
Built on top of it, \unifiedmodel{} eliminates the need for task-specific visual representations, and unifies diverse vision-language tasks under a single autoregrsssive prediction paradigm. 
By formulating both understanding and generation as next-token prediction in a shared continuous space, it seamlessly supports multi-round, in-context tasks such as iterative understanding, generation and editing.
Empirically, we find that using a unified continuous visual representation reconciles the competing requirements on the tokenizers by the understanding and generation tasks, thereby leading to state-of-the-art level performance across both domains.
We hope our findings will facilitate unified visual tokenization in the continuous domain. 
Inference code and model weights are released to benefit community.
}}

\date{Oct 7, 2025\vspace{-1mm}}
\gtechdata[\github]{\url{https://github.com/inclusionAI/Ming-UniVision}} 
\gtechdata[\huggingface]{\url{https://huggingface.co/inclusionAI}} 
\gtechdata[\modelscope]{\url{https://www.modelscope.cn/organization/inclusionAI}} 
\begin{document}
\maketitle

\section{Introduction}
\label{sec:intro}

\begin{figure}[p]
\centering
\includegraphics[width=\linewidth]{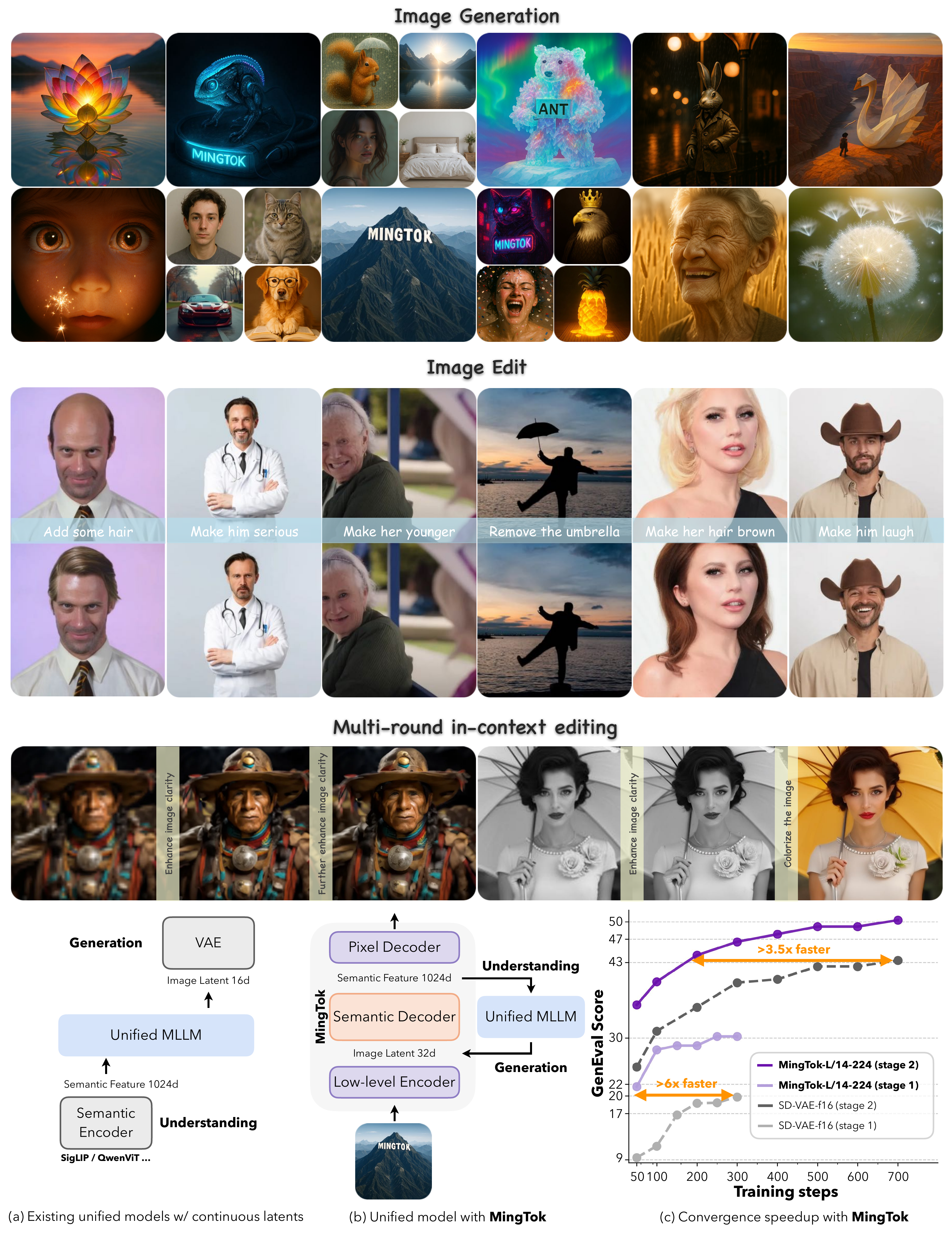}
\caption{\textbf{Conceptual comparison and qualitative examples of \model{}}. (a) Existing models using continuous latent spaces for unified visual understanding and generation uses two sets of representations for visual contents. (b) \model{} employes a unified tokenizer for generating semantic and low-level image representatinos. (c) Compared with SD-VAE~\citep{sd_2022}, \model{} achieves over 3.5 times acceleration for text-to-image generation. }
\label{fig:intro_image}
\end{figure}

Autoregression has emerged as a powerful and general technique to model diverse data modalities, achieving remarkable success in language~\citep{gpt3,dubey2024llama3}, audio~\citep{chu2023qwenaudio,ding2025kimiaudio}, and omni-modal sequences~\citep{hurst2024gpt4o,ai2025mingomni}.
By formulating both visual understanding and image generation as sequential prediction problems, autoregressive models have achieved competitive results in vision-language understanding~\citep{bai2025qwen25vl,guo2025seed15vl} and generation~\citep{li2024mar,tian2024var}.
This has motivated recent efforts to unify vision-language understanding and generation within a single autoregressive framework~\citep{ai2025minguni,pan2025metaquery,deng2025bagel}, to fully leverage in-context learning, compositional reasoning, and task generalization abilities inherent in large language models.

In terms of input and output, visual understanding and generation are inverse tasks in some sense, where one 
where one maps pixels to semantic concepts and the other synthesizes pixels from textual descriptions. 
Most of current multi-modal models handle them as two separate tasks with totally distinct approaches~\citep{wu2025qwenimage,pan2025metaquery,deng2025bagel}, as they typically use distinct representation spaces for understanding and generation.
This discrepancy stems from the fundamentally different and competing requirements for their underlying representations,
where understanding tasks favor high-dimentional semantic features~\citep{radford2021clip,zhai2023siglip}, while generation demands compact and low-dimentional structured latent codes that preserve fine-grained visual details~\citep{esser2021sd,sd_2022}. As a result, most existing frameworks resort to asymmetric designs~\citep{deng2025bagel,fan2025unifluid,chen2025januspro}, using distinct tokenization schemes for the two tasks, which introduces optimization difficulty and architectural complexity to the unified system.

Recent efforts have sought to bridge this gap through shared visual tokenizers~\citep{qu2025tokenflow,wu2024vilau,jiao2025unitoken}. However, these approaches often rely on discrete latent representations, which introduces quantization error and limits representation capacity. This degradation not only constrains the fidelity of generated images, but also impairs semantic expressiveness, leading to suboptimal performance on understanding tasks~\citep{ma2025unitok}.

To address these limitations, we present \model{}, a visual tokenizer with a continuous latent space that enables unified autoregressive vision-language understanding and generation.
The core of \model{} is a three-stage sequential architecture designed to reconcile the competing representational demands of understanding and generation:
\textbf{(i)}~A \textit{low-level encoder} maps input images into a compact, continuous latent representation optimized for efficient autoregressive generation;
\textbf{(ii)}~A \textit{semantic decoder} progressively expands this compact latent sequence into high-dimensional semantic features through autoregressive refinement, producing expressive high-dimensional representations suitable for vision-language reasoning;
\textbf{(iii)}~A \textit{pixel decoder} reconstructs the original image from the semantic features, ensuring high-fidelity image reconstruction.
The entire model is end-to-end optimized with a masked image modeling objective that imposes supervision over both intermediate representations and reconstructed pixels, thereby simultaneously enhancing semantic richness and shaping the compact latent space to be more conducive to autoregressive generation.

On top of \model{}, we build a unified multi-modal model, \unifiedmodel{}, that adopts a single autoregressive architecture for joint text and image modeling. 
Both understanding and generation operate on a shared high-level semantic space as input, while generating compact continuous latents in a token-by-token manner, resulting in significantly reduced architectural complexity compared to existing dual-encoder designs, as illustrated in Fig.~\ref{fig:intro_image}.
This unified paradigm enables strong performance across both vision-language understanding and text-to-image generation, demonstrating that a single autoregressive framework can support diverse capabilities through universal visual representations.

Crucially, by processing both modalities within a shared next-token prediction paradigm and maintaining unified representations within each modality, our model establishes a coherent interface for multimodal sequence modeling.
This enables seamless integration: any prefix sequence, whether composed of text tokens, visual latents, or mixed modalities, can be directly consumed to condition downstream generation or reasoning in either modality.
As a result, the same architecture supports not only text-to-image and image-to-text tasks, but also unlocks complex multi-round interactions, including iterative super-resolution, region-aware editing, and sequential refinement, \textit{e.g.}, upsample $\rightarrow$ colorize (Fig.~\ref{fig:qual-photo-restoration-super-res}) and segment $\rightarrow$ edit (Fig.~\ref{fig:got_process}), all within a single and coherent framework. These results suggest that unified modeling through persistent latent representations opens up new possibilities for interactive, human-in-the-loop vision systems, where generation, editing, and understanding are no longer isolated pipelines, but interwoven steps in a continuous visual dialogue. We believe this shift toward dynamic, context-aware multimodal interaction will enable more flexible, intuitive, and cognitively aligned models in the future.

Our contributions are as follows:
\begin{itemize}
    \item We propose \model{}, a continuous visual tokenizer that unifies generation and understanding through a three-stage architecture, eliminating quantization error while supporting both compact latents and rich semantics.
    
    \item Built on \model{}, we introduce \unifiedmodel{}, a unified autoregressive framework where diverse vision-language tasks are cast as next-token prediction, enabling seamless integration of perception, generation, and editing without task-specific representations.
    
    \item We demonstrate efficient multi-round in-context editing, enabled by unified latent representations that eliminate the need for repeated encoding through distinct tokenizers. By drastically reducing the visual token count—requiring up to 66\% fewer input tokens compared to prior unified architectures—\unifiedmodel{} enables faster iteration and lower memory overhead in interactive generation workflows.

\end{itemize}
\section{Unified Visual Tokenization without Vector Quantization}
\label{sec:mingtok-arch}

\begin{figure}[t]
\centering
\includegraphics[width=\textwidth]{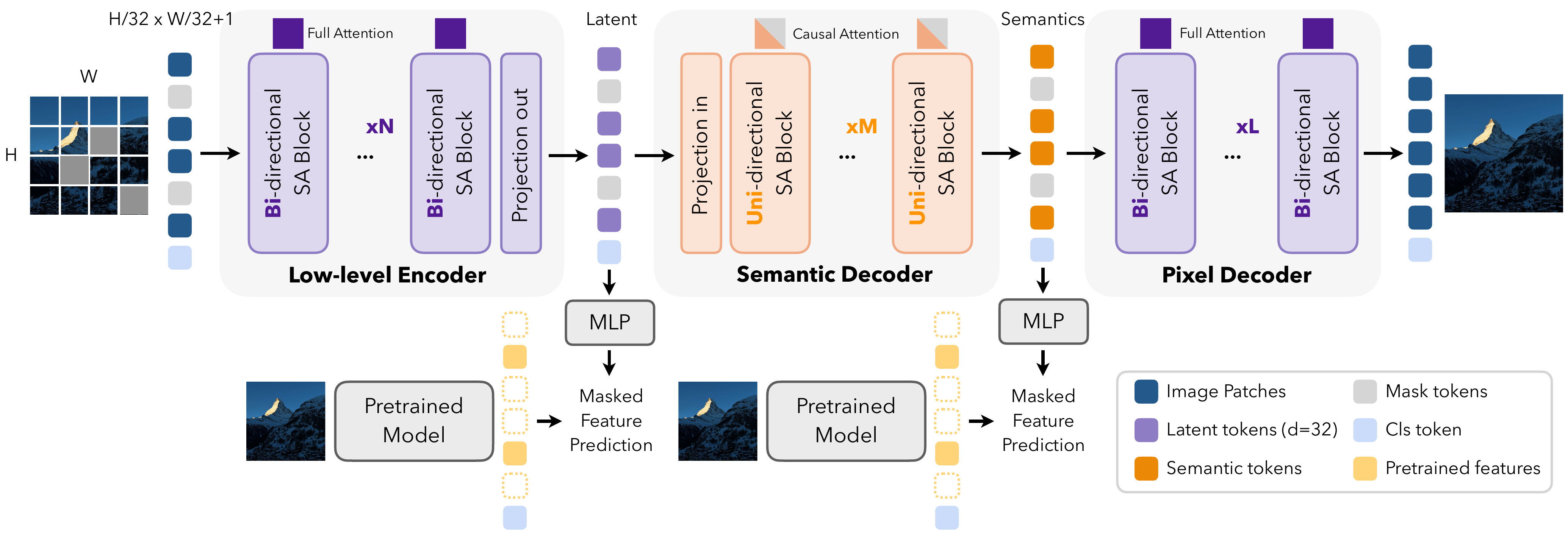}
\caption{
The model architecture and the training objectives of \model{}. \model{} performs image compression, semantic decoding and image reconstruction sequentially through low-level encoder, semantic decoder, and pixel decoder. 
During training, both the image latent and the semantic features are supervised by pre-trained visual encoders with masked feature prediction, while the pixel decoder is trained by masked and unmasked image reconstruction. 
}
\label{fig:mingtok_arch}
\end{figure}

\subsection{Tokenizer Model Architecture}

A unified visual tokenizer must produce compact and high-dimensional patch representations, trained for both understanding and generation. As shown in Fig.~\ref{fig:mingtok_arch}, \model{} employs a three-stage sequential architecture composed of a low-level encoder, a semantic decoder, and a pixel decoder, all primarily implemented using vision transformer blocks with different masking strategies. 

\textbf{Low-level encoder} transforms raw image pixels into compact latent embeddings for autoregressive generation. 
The compression ratio of the image latents is entirely determined by the patch embedding layer. The number of tokens are kept the same throughout the low-level encoder and the semantic decoder. 
A linear projection layer equipped with a channel-averaging shortcut serves as the output layer, compressing the latent channel dimension to a generation-friendly size, typically 16 or 32. 
Within the low-level encoder, full attention is employed to holistically model spatial dependencies and structural patterns in the input images.

\textbf{Semantic decoder} is connected after the low-level encoder, expanding the compact latent embeddings into rich semantic features for visual understanding. A linear projection layer with a channel-repeating shortcut serves as the input layer, expanding the latent channel dimension for semantic understanding. In order to support per-token autoregressive generation, the semantic decoder adopts causal attention. 

\textbf{Pixel decoder} reconstructs the original images from the high-dimentional semantic features generated by the semantic decoder, serving as the final rendering stage in both reconstruction and autoregressive generation. To recover the fine visual details, we apply a pixel unshuffle layer~\citep{shi2016pixelshuffle} before the transformer blocks in the pixel decoder to increase the number of visual tokens and reduce the effective patch size. We find that this leads to significantly better texture fidelity and edge sharpness in the output images. Within the transformer layers, full self-attention is used to capture potential long-range dependencies across the entire feature map.

The architectural hyperparameters are shown in Table.~\ref{tab:mingtok_hyperparam}.

\begin{table}[t]
    \centering
    \small
    \begin{tabular}{lcccccccc}
    \toprule
         \textbf{Component} & \textbf{Resolution} & \textbf{Patch size}  & \textbf{Head dim} & \textbf{Num heads} & \textbf{Embed dim} & \textbf{Depth} & \textbf{Out dim} & \textbf{Attention}\\
          \midrule
         Low-level Encoder & 512 & 32 & 64 & 12 & 768 & 12 & 32 & Full \\
         Semantic decoder & 512 & 32 & 64 & 16 & 1024 & 24 & 1024 & Causal \\
         Pixel decoder & 512 & 16 & 64 & 16 & 1024 & 24 & 3 & Full \\
    \bottomrule
    \end{tabular}
    \caption{The architectural hyperparameters of \model{}.}
    \label{tab:mingtok_hyperparam}
\end{table}

\subsection{Training}

The training of \model{} is guided by three key design principles:
\textbf{(i)}~the generative latent space should be both \textit{structured} and \textit{compact} to enable fast convergence and efficient autoregressive modeling;
\textbf{(ii)}~the semantic space is required to support strong scalability and generality in vision-language understanding tasks;
\textbf{(iii)}~the model should be able to achieve high-fidelity image reconstruction to ensure the quality of the generated visual contents.
Therefore, we adopt a multi-task learning framework built upon the masked image modeling paradigm~\citep{wei2022maskefeat,xie2022simmim,fang2024eva02,he2022mae}, with three complementary objectives targeting different components of the architecture.

\textbf{Structured latent space regularization.}
As shown in prior work on variational autoencoders in visual generative models~\citep{yao2025vavae,chen2025maetok,skorokhodov2025diffusability,chen2024dcae,chen2025hieratok}, the structure and compactness of the latent space critically influence generation efficiency and downstream model convergence. To encourage a highly structured and compressed representation, we use a spatial compression ratio of 32.
The low-level encoder and latent space are trained via \textit{masked feature prediction}.
The token sequences are randomly masked at the input side of the low-level encoder, and the masked tokens at the output side is used to predict the corresponding features from a pre-trained vision foundation model (\textit{e.g.}, DINOv2~\citep{oquab2023dinov2}) at the same spatial locations.
This objective regularizes the latent space with rich semantic and structural priors required for autoregressive visual generation, and our empirical verification demonstrate the effectiveness of such a training objective.

\textbf{Scalable semantic representation learning.}
To build a semantic decoder that supports strong generalization and scaling in multimodal reasoning, we apply the same masked feature prediction paradigm—known for its superior scalability over plain feature distillation~\citep{fang2023eva}. 
Specifically, the compact latent sequence from the low-level encoder with partial token masked is further passed through the semantic decoder, which is autoregressively expanded to high-dimensional semantic feature sequence. 
The expanded semantic features at the masked locations are supervised by the feature representations of the visual backbones aligned with text semantics during pre-training (\textit{e.g.,} CLIP~\citep{radford2021clip}). 
This ensures that the semantic space develops expressive, scalable visual representations suitable for complex vision-language tasks. 

\textbf{Pixel Reconstruction Objective.}
To enhance the robustness and fidelity of the pixel decoder, we train it under both \textit{masked} and \textit{unmasked} conditions, where the representations of both observed patches and masked patches are received by the pixel decoder. 
The decoder then learns to reconstruct the full image.
This dual-setting supervision forces the decoder to recover fine-grained details even when some latents are missing or noisy—mimicking the autoregressive generation process where tokens are generated sequentially. As a result, the decoder becomes more robust and capable of producing high-quality reconstructions.

Since the pixel decoder reconstructs image from semantic features produced by the semantic decoder, it also requires the semantic decoder to preserve sufficient low-level details, which in turn enhances the fine-grained visual perception capability of the semantic features. In our early experiments, we found adding reconstruction objective to masked feature prediction can notably improve the downstream understanding performance when combined with multi-modal LLMs. 

\section{Unified Image Understanding and Generation}
\subsection{Multimodal Model Architecture}

\begin{figure}[t]
\centering
\includegraphics[width=0.95\textwidth]{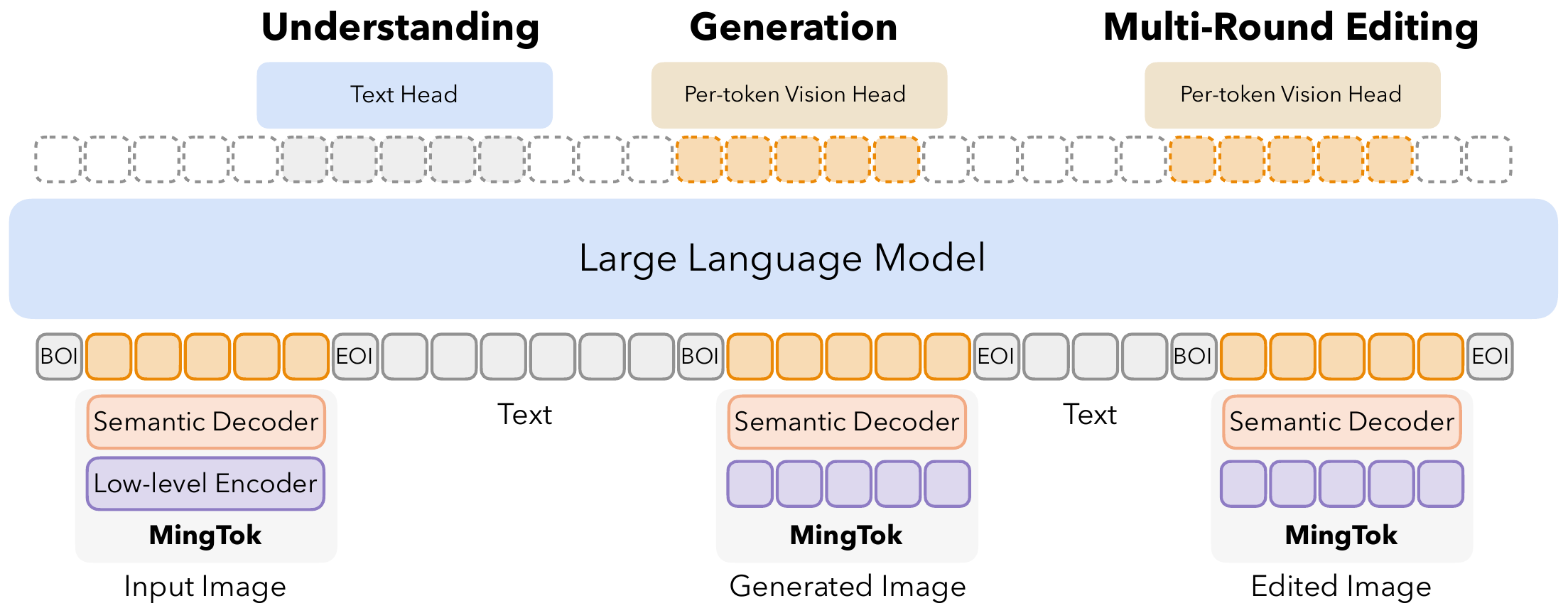}
\caption{The architecture of \unifiedmodel{}. 
Owing to the autoregressive semantic decoding capability of \model{}, both image understanding (image-to-text generation) and image synthesis (text-to-image generation) can be formulated consistently with the same next-token prediction paradigm and unified input representation space. 
This allows our unified multimodal model to support multi-round in-context tasks, seamlessly switch from understanding to generation/editing task, and vice versa. 
}
\label{fig:unified_gen_arch}
\end{figure}

The overall architecture of \unifiedmodel{}, a multi-modal model for image understanding, generation, and in-context multi-round editing is shown in Fig.~\ref{fig:unified_gen_arch}.
Our unified model achieves seamless integration of vision and language through two key unifications enabled by \model{}:

\textbf{Unified Visual Input Representation.}
For both understanding and generation tasks, the language model consistently receives high-level semantic features produced by the semantic decoder. 
In image understanding, the representation is derived from real images. 
The input is first encoded into compact continuous latents via the low-level encoder, then passed through the semantic decoder to produce rich, text-aligned visual embeddings. Since the entire image is available upfront, all semantic tokens are computed in parallel.
In autoregressive image generation, instead of encoding an observed image, the vision head of the language model generates compact latents one token at a time. Each generated latent token is immediately expanded into its corresponding semantic feature by the semantic decoder, which is then fed into the language model as the contextual input for next token prediction. 
This ensures a unified interface for multimodal interaction, regardless of whether the visual content is perceived or synthesized. 

\textbf{Unified Next-Token Prediction.}
On the output side, both modalities are generated autoregressively under a shared sequence modeling paradigm. Textual tokens are predicted using the standard language model head, preserving full compatibility with pre-trained LLMs. For visual content, a per-token vision head is attached to the language model to predict compact continuous latents one patch at a time, enabling seamless interleaving of text and image generation within the same autoregressive framework.

This vision head is inspired by recent works on unified multi-modal modeling~\citep{li2024mar, fan2025unifluid}, but incorporates two key structural improvements. First, we replace the diffusion-based denoising head with a rectified flow~\citep{liu2022flow} prediction objective, which allows for faster convergence and fewer inference steps. Second, we adopt a SwiGLU-based feed-forward network (FFN)~\citep{shazeer2020swiglu} in place of the standard MLP block, which we find empirically improves latent prediction accuracy and final image quality under the same parameter budget.

Together, the unified input representation and next-token prediction enable a single model to universally handle understanding, generation, and editing, simplifying the architecture for multi-round in-context image understanding, generation, and manipulation, which we elaborate in Sec.~\ref{sec:multi-round}.

\subsection{Multi-round In-context Image Understanding, Generation and Manipulation}
\label{sec:multi-round}

While recent unified vision-language models have made significant advances in joint understanding and generation~\citep{deng2025bagel,chen2025januspro,fan2025unifluid}, they still follow a fragmented generate-then-understand pipeline. Whether through dual branches~\citep{deng2025bagel,shi2024lmfusion} or separate tokenizers~\citep{fan2025unifluid}, these approaches decouple the generation and perception processes into distinct spaces, leading to repeated encoding and decoding cycles, imposing significant overhead on iterative in-context refinement. There are several major obstacles preventing such architectures from supporting efficient and scalable \emph{multi-round in-context image generation}:

\textbf{DiT is structurally incapable of multi-round in-context editing. }
Diffusion Transformers (DiTs), such as \textit{FLUX.1}, are architecturally designed to generate a fixed number of images per forward pass. 
During training, the diffusion transformers are configured to generate images based on a pre-determined number of reference images, resulting in a static input-output structure.
This rigidity limits their ability to dynamically extend generation sequences or flexibly interleave image editing steps within a single context, rendering them ill-suited for adaptive, multi-round tasks.

\textbf{Hybrid AR–Diffusion is burdened with dual-branch overhead. }
Hybrid AR-Diffusion designs~\citep{deng2025bagel,shi2024lmfusion} integrate autoregressive and per-image diffusion to enable in-context multi-round generation and editing. 
While the dual-branch architecture supports multi-round generation, it introduces significant computational and implementation overhead:
\begin{itemize}
    \item \textbf{Training computation overhead:}  
    Hybrid models maintain multiple distinct representations per image, \textit{i.e.,} semantic features for understanding, noised latents for denoising, and clean latents for conditioning future steps. This substantially increases the effective token sequence length during training, leading to higher memory consumption and longer training times.
    \item \textbf{Training complexity: }Unconventional attention masking schemes are required to manage cross-feature space and cross-round dependencies:
    \begin{itemize}
        \item Noisy tokens from prior generation rounds are masked out in subsequent generation steps, ensuring that only the clean latents are observed in future image generation processes.
        \item Different masking strategies are applied to across distinct feature spaces: \textit{Causal attention} over semantic features and \textit{full attention} over image latents to support global denoising.
    \end{itemize}
    \item \textbf{Inference inefficiency:}  
    Multi-round generation requires frequent conversions between heterogeneous spaces:  
    \textit{latent space} (generation) $\rightarrow$ \textit{pixel space} (full decoding via VAE) $\rightarrow$ \textit{feature space} (semantic encoding via understanding encoder). 
    After each round of generation, a full decode-encode cycle is needed, increasing both latency and computational overhead.
\end{itemize}
Consequently, while AR-Diffusion frameworks supports multi-round editing, they suffer from increased architectural complexity, reduced training stability, and inefficient inference, posing practical challenges for scalable deployment.

\textbf{Unified AR is limited by separated tokenization. }  
Architectures such as \textit{UniFluid}~\citep{fan2025unifluid} adopt a single-branch autoregressive loop to unify understanding, generation and editing within a shared sequence modeling framework. 
Compared to AR-Diffusion hybrid models, their unified architectures simplify training and inference by relying on a single next-token prediction objective, eliminating the need for complex masking schemes.

However, they still rely on distinct representations for understanding and gneration, which necessitates frequent conversions between domains during multi-round editing.
Moreover, during training, both semantic and generative token sequences are processed in parallel, effectively doubling the input length and increasing memory and computational overhead.
As a result, despite architectural simplification, unified autoregressive models still inherit key inefficiencies from hybrid approaches, particularly in terms of latency and scalability in iterative editing scenarios.

\begin{figure}[t]
\centering
\includegraphics[width=0.95\textwidth]{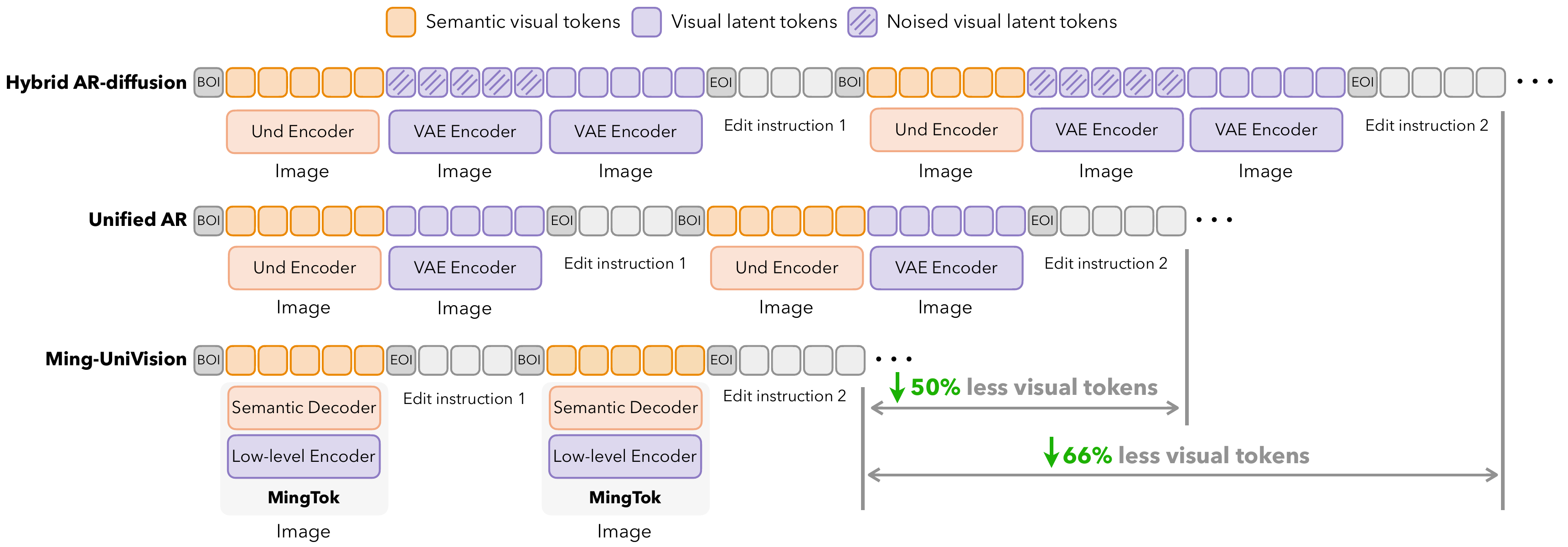}
\caption{
Comparison of input token structures across different unified model architectures. 
\unifiedmodel{} reduces the number of input visual tokens by 66\% compared to hybrid AR-diffusion models~\citep{shi2024lmfusion,deng2025bagel} and by 50\% compared to existing unified autoregressive models~\citep{fan2025unifluid}, thanks to the unified representation enabled by \model{}.
}
\label{fig:unified_gen_incontext}
\end{figure}

In contrast, \unifiedmodel{} unifies understanding and generation within a \textit{single continuous token space} (see Section~\ref{sec:mingtok-arch}).
This is made possible by the unified input representation enabled by \model{}, which allows the high-dimensional features from the \textit{semantic decoder} to be reused as conditional inputs for generation or editing, without costly pixel-space detour.
As illustrated in Fig.~\ref{fig:unified_gen_incontext}, this design supports efficient \textit{in-context} interaction, enabling reversible edits, faithful reconstruction, and iterative refinement, all while preserving the full context in the latent space.

This design enables a seamless, in-place iterative workflow during inference: after generating an image, its semantic feature representation \(S_t\) remains in the latent space. For the next step \(t+1\), \(S_t\) is concatenated with a new textual instruction (e.g., ``add a hat'') and fed back into the model to produce an updated semantic feature \(S_{t+1}\). Because the entire process is executed purely in the latent space---bypassing costly pixel-space re-encoding---it avoids cumulative quality degradation, preserves visual fidelity, and supports low-latency, multi-round editing and generation that can fluidly interleave image understanding and free-form content creation.

\subsection{Training}

The training of our unified multi-modal model is divided into two main phases: \textit{pre-training} on large-scale web data to establish foundational vision-language capabilities, followed by \textit{supervised fine-tuning (SFT)} on curated datasets to enhance instruction-following behavior and support complex multimodal interactions. This staged strategy ensures stable optimization while enabling progressive acquisition of understanding, generation, and editing skills.

\subsubsection{Pre-training}
Pre-training consists of two stages designed to align representations before joint modeling.

\textbf{Stage 1: MLP and rectified flow head warm-up. }In this initialization phase, we focus on training the MLP between \model{} and the LLM, and the per-token vision head for latent prediction. The \model{} and the LLM backbone remain fixed during this stage. We use a mix of image-text pairs, with approximately 30\% dedicated to understanding tasks and 70\% to autoregressive generation tasks. This warms up both the vision-to-language and the language-to-vision pathways.

\textbf{Stage 2: Joint image understanding and generation pre-training. }In this stage, we aim to build robost single-turn vision-language capabilities using large-scale image-text data. Since the prediction of per-token rectified flow head is limited to the current token, it relies on the language model to model sequential relationships among visual tokens. Hence, in this stage, we unlock the language model, allowing it to capture inter-token structure during autoregressive generation. 

To enhance fine-grained visual perception without destabilizing the pre-trained latent space, we introduce \textit{mixed resolution training} and selectively unlock only the semantic decoder of \model{}, keeping the low-level encoder fixed. During understanding tasks, images are resized to 1024$\times$1024, and the semantic decoder learns to produce high-fidelity, detail-rich embeddings taht align with text semantics. For text-to-image generation, the inputs remain at 512$\times$512 in consideration of the computational efficiency and compatibility with the pre-trained compact latent space.

This setup enables the model to perceive fine details during understanding while retaining stable, fast generation—critical for downstream editing and in-context interaction. The training data consists of approximately 25\% image-text understanding pairs, 70\% text-to-image generation samples, and 5\% general NLP tasks.

\subsubsection{Supervised Finetuning}

After pre-training, we perform supervised fine-tuning on high-quality datasets and custom instruction-following corpora to enhance instruction adherence, generation fidelity, and support for complex multi-modal interactions. This phase has two stages that progressively introduce task complexity. 

\textbf{Stage 1: Image understanding and generation.} This stage focuses on aligning the model with human intent in standard vision-language tasks. We freeze \model{} and unlock the remaining parts, as we observe no performance gain when the semantic decoder is unlocked during this stage. Mixed resolution training is further employed in this stage. The data distribution includes approximately 30\% for understanding tasks, 10\% for NLP tasks, and 60\% for text-to-image generation. 

\textbf{Stage 2: Image understanding, generation, and in-context image manipulation.} To enable multi-round context-aware instructions such as iterative editing and refinement, we introduce a final fine-tuning stage focused on image generation and in-context manipulation. We constructed the instruction chains for the model to learn in-context image manipulation (see details in Sec.~\ref{sec:multiround-eval}). The training strategy follows the same strategy as stage 1, with data composition significantly shifted, 15\% understanding, 5\% NLP, 35\% standard text-to-image generation, and 55\% single or multi-round editing tasks. 

\section{Evaluations}

\unifiedmodel{} is evaluated on a wide range of image understanding and generation benchmarks. For the comparison with existing approaches, we employ Ling-lite language model featuring 2.8 billion activated parameters.

\subsection{Multi-modal Understanding}

\begin{table}[t]
    \centering
    \setlength{\tabcolsep}{3pt}
    \renewcommand{\arraystretch}{1.2}
    \scriptsize
    \caption{\textbf{Quantitative evaluations} on MMBench~\citep{liu2024mmbench}, MMStar~\citep{chen2024mmstar}, MMMU~\citep{yue2024mmmu}, MathVista~\citep{lu2023mathvista}, HallusionBench~\citep{guan2024hallusionbench}, AI2D~\citep{kembhavi2016ai2d}, MM-Vet~\citep{yu2023mmvet}, OCRBench~\citep{liu2024ocrbench}, and MME~\citep{fu2023mme}.
    }
    \label{sota_result_understanding}
    \scalebox{1.0}{
    \begin{tabular}{lccccccccc}
        \toprule
        \textbf{Model} & \textbf{MMB$ \uparrow$} & \textbf{MMS$ \uparrow$} & \textbf{MMMU$ \uparrow$} & \textbf{MathV$ \uparrow$}  & \textbf{Hall$ \uparrow$}  & \textbf{AI2D$ \uparrow$}  & \textbf{MM-Vet$ \uparrow$} & \textbf{OCRBench$ \uparrow$}& \textbf{MME$ \uparrow$}\\
        \midrule
        \multicolumn{9}{c}{\textit{Understanding Only}}  \\
        Emu$3$-Chat~\citep{wang2024emu3} & 58.5 & - & 31.6 & - & - & - & 37.2 & 687 & -\\
        Qwen2.5-VL-3B~\citep{bai2025qwen25vl} & 79.1 & 55.9 & 53.1 & 62.3 & 46.3 & 81.6 & - & 797 & 2157 \\
        Qwen2.5-VL-7B~\citep{bai2025qwen25vl}   & 83.5 & 63.9 & 58.6 & 68.2 & 52.9 & 83.9 & 67.1 & 864 & 2347 \\
        InternVL2.5-4B~\citep{chen2024internvl25} & 81.1 & 58.3 & 52.3 & 60.5 & 46.3 & 81.4 & 60.6 & 828 & 2338 \\
        InternVL2.5-8B~\citep{chen2024internvl25}  & 84.6 & 62.8 & 56.0 & 64.4 & 50.1 & 84.5 & 62.8 & 822 & 2344\\
        DeepSeek-VL2~\citep{wu2024deepseekvl2}  & 79.6 & 61.3 & 51.1 & 62.8 & - & 81.4 & - & 811 &  2253 \\   
        \midrule
        \multicolumn{9}{c}{\textit{Unified model, Separate representation}}\\
         Janus-Pro-7B~\citep{chen2025januspro}  & 79.2 & - & 41.0 & - &  - & - & 50.0 & - & -\\
         LMFusion~\citep{shi2024lmfusion} & - & - & 41.7 & - & - & - & - & - & 1603 \\
         MetaQuery-L~\citep{pan2025metaquery} & 78.6 & - & 53.1 & - & - & - & 63.2 & - & -\\
         Show-o2-7B~\citep{xie2025showo2} & 79.3 & 56.6 & 48.9 & - & - & 78.6 & - & - & - \\
         BLIP3-o 4B~\citep{chen2025blip3o} & 78.6 & - & 46.6 & - & - & - & 60.1 & - & 2161 \\
         BAGEL~\citep{deng2025bagel} & 85.0 & - & 55.3 & 73.1 & - & - & 67.2 & - & 2388 \\
        \midrule
        \multicolumn{9}{c}{\textit{Unified model, Unified representation}} \\
         VILA-U~\citep{wu2024vilau} & - & - & - & - & - & - & 33.5 & -& 1402\\
         TokenFlow-XL~\citep{qu2025tokenflow}  & 76.8 & - & 43.2 & - & - & - & 48.2& - & 1922\\
         UniTok~\citep{ma2025unitok} & - & - & - & - & - & - & 33.9 & - & 1448 \\
         Harmon-1.5B~\citep{wu2025harmon} & 65.5 & - & 38.9 & - & - & - & - & - & 1476 \\
         TokLIP~\citep{lin2025toklip} & 67.6 & - & 43.1 & - & - & - & 29.8 & - & - \\
        \cdashline{1-10}
        \textbf{\unifiedmodel{}-16B-A3B (Ours)} & 78.5 & 63.7 & 40.3 & 66.6 & 47.8 & 82.8  & 64.2 & 724 & 2023 \\
        \bottomrule
    \end{tabular}
    }
    \label{tab:sota_comparison_und}
\end{table}

\begin{table}[t]
    \centering
    \setlength{\tabcolsep}{2.0pt}
    \renewcommand{\arraystretch}{1.2}
    \scriptsize
    \caption{\textbf{Evaluation of text-to-image generation ability on GenEval}~\citep{ghosh2023geneval} \textbf{and DPG-Bench}~\citep{hu2024ella}. $\dagger$ denotes performance obtained by rewritten prompts.
    }
    \vspace{-2mm}
    \scalebox{1.0}{
    \begin{tabular}{lccccccccc}
        \toprule
        \textbf{Method}  & \textbf{Single Obj.$\uparrow$} & \textbf{Two Obj.$\uparrow$} & \textbf{Counting$\uparrow$} & \textbf{Colors$\uparrow$} & \textbf{Position$\uparrow$} & \textbf{Color Attri.$\uparrow$} & \textbf{Overall$\uparrow$} & \textbf{DPG-Bench$\uparrow$} \\
        \midrule
        \multicolumn{8}{c}{\textit{Generation Only}}  \\
        LlamaGen~\citep{sun2024autoregressive}  & 0.71 & 0.34 & 0.21 & 0.58 & 0.07 & 0.04 & 0.32 & -\\
        PixArt-$\alpha$~\citep{chen2023pixart} &  0.98 & 0.50 & 0.44 & 0.80 & 0.08 & 0.07 & 0.48 & - \\
        SDv$2.1$~\citep{sd_2022} &  0.98 & 0.51 & 0.44 & 0.85 & 0.07 & 0.17 & 0.50 & -\\
        DALL-E $2$~\citep{ramesh2022hierarchical}  & 0.94 & 0.66 & 0.49 & 0.77 & 0.10 & 0.19 & 0.52 & -\\
        Emu$3$-Gen~\citep{wang2024emu3}  & 0.98 & 0.71 & 0.34 & 0.81 & 0.17 & 0.21 & 0.54 & 80.60\\
        SDXL~\citep{podell2023sdxl} &  0.98 & 0.74 & 0.39 & 0.85 & 0.15 & 0.23 & 0.55 & 74.65\\
        DALL-E $3$~\citep{dalle3}  & 0.96 & 0.87 & 0.47 & 0.83 & 0.43 & 0.45 & 0.67 & 83.50\\
        SD3-Medium~\citep{esser2024scalingrectifiedflowtransformers} & 0.99 & \textbf{0.94} & 0.72 & 0.89 & 0.33 & 0.60 & 0.74 & 84.08\\
        \midrule
        \multicolumn{8}{c}{\textit{Unified model, Separate representation}}\\
        Show-o~\citep{xie2024showo} &  0.95 & 0.52 & 0.49 & 0.82 & 0.11 & 0.28 & 0.53 & -\\
        Ming-Lite-Uni~\citep{ai2025minguni} & 0.99 & 0.76 & 0.53 & 0.87 & 0.26 & 0.30 & 0.62 & -\\
        Janus-Pro-1B~\citep{chen2025januspro} &  0.98 & 0.82 & 0.51 & 0.89 & 0.65 & 0.56 & 0.73 & 82.63 \\
        Janus-Pro-7B~\citep{chen2025januspro} & 0.99 & 0.89 & 0.59 & 0.90 & 0.79 & 0.66 & 0.80 & 84.19 \\
        Show-o2-7B~\citep{xie2025showo2} & \textbf{1.00} & 0.87 & 0.58 & 0.92 & 0.52 & 0.62 & 0.76 & \textbf{86.14} \\
        MetaQuery-L$\dagger$~\citep{pan2025metaquery} &  - & - & - & - & - & - & 0.78 & 81.10 \\
        Blip3-o 4B~\citep{chen2025blip3o} &  - & - & - & - & - & - & 0.81 & 79.36 \\
        BAGEL~\citep{deng2025bagel} & 0.99 & \textbf{0.94} & \textbf{0.81} & 0.88 & 0.64 & 0.63 & 0.82 & - \\
        \midrule
        \multicolumn{8}{c}{\textit{Unified model, Unified representation}} \\
        Harmon-1.5B~\citep{wu2025harmon} & 0.99 & 0.86 & 0.66 & 0.85 & 0.74 & 0.48 & 0.79 & - \\
        TokenFlow-XL~\citep{qu2025tokenflow} &  0.95 & 0.60 & 0.41 & 0.81 & 0.16 & 0.24 & 0.55 & 73.38 \\
        \textbf{\unifiedmodel{}-16B-A3B (Ours)} & \textbf{1.00} & 0.93 & 0.59 & \textbf{0.93} & \textbf{0.92} & \textbf{0.70} &  \textbf{0.85} & 82.12 \\
        \bottomrule
    \end{tabular}
}
    \label{tab:geneval}
\vspace{-2mm}
\end{table}

The understanding capability of our model is summarized in Table~\ref{tab:sota_comparison_und}, where we compare with existing models of comparable parameter scale. Compared to both dedicated vision-language understanding models and unified architectures that employ separate representations for perception and generation, our model achieves comparable overall performance. 

Notably, \unifiedmodel{} shows competitive results on MMStar~\citep{chen2024mmstar}, HallusionBench~\citep{guan2024hallusionbench}, AI2D~\citep{kembhavi2016ai2d}, and MM-Vet~\citep{yu2023mmvet}, which evaluate semantic reasoning and hallucination detection. This indicates that  the shared semantic representation learned by \model{} is sufficiently expressive for general-purpose vision-language understanding. However, we observe a performance gap on OCRBench~\citep{liu2024ocrbench} and MMMU~\citep{yue2024mmmu}, where fine-grained recognition is critical. This suggests limitations in preserving character-level details, which is likely due to the compressed nature of the latent space used during autoregressive generation as well as the causal architecture of the semantic decoder. 
We leave the optimization for detail-preserving latents and semantic representations for future work.

\subsection{Visual Generation}
The visual generation capabilities of \unifiedmodel{} are benchmarked in Table \ref{tab:geneval}, where we evaluate on GenEval~\citep{ghosh2023geneval} and DPG-Bench~\citep{hu2024ella}. Compared with existing models, our model achieves state-of-the-art performance on the overall GenEval benchmark.

Notably, it particularly excels in attribute control and spatial reasoning, outperforming all other models in the \textbf{Position} (0.92), \textbf{Colors} (0.93), and \textbf{Color Attribute} (0.70) sub-tasks. The significant lead in the position-related tasks underscores our model's superior compositional control. This strong performance, combined with faster training convergence inherent to our unified architecture, highlights the effectiveness of the shared semantic space in guiding image synthesis.

We attribute these improvements to the joint perception–generation representation, which facilitates both semantic grounding and efficient optimization. Further enhancements in fine-grained detail generation remain an avenue for future work.

\subsection{Visual Editing}

\begin{table*}[t]
  \centering
  \captionsetup{justification=centering}
  \begin{minipage}[t]{0.55\textwidth}
    \scriptsize
    \captionsetup{justification=raggedright, singlelinecheck=false}
    \caption{\textbf{Evaluation on the image reconstruction ability} on the validation set of ImageNet~\citep{deng2009imagenet}.}
    \centering
    \setlength{\tabcolsep}{2pt}
    \begin{tabular}{lcccccc}
      \toprule
      \textbf{Tokenizer} & \textbf{Res.} & \textbf{\# Tokens} & \textbf{rFID$ \downarrow$} & \textbf{PSNR $\uparrow$} & \textbf{SSIM $\uparrow$} & \textbf{LPIPS $\downarrow$}\\
      \midrule
      \multicolumn{7}{l}{\textit{Specialized tokenizers}} \\
      SD-VAE~\citep{esser2024scalingrectifiedflowtransformers} & 256 & 1024 & 1.06 & 28.62 & 0.86 & - \\
      GigaTok~\citep{xiong2025gigatok} & 256 & 256 & 0.51 & 21.32 & 0.69 & 0.21 \\
      VA-VAE~\citep{yao2025vavae} & 256 & 256 & 0.26 & 28.59 & 0.80 & 0.09 \\
      DC-AE~\citep{chen2024dcae} & 512 & 64 & 0.22 & 26.15 & 0.71 & 0.08 \\
      MAE-Tok~\citep{chen2025maetok} & 512 & 128 & 0.62 & - & - & - \\
      TexTok~\citep{zha2025textok} & 512 & 256 & 0.73 & 24.45 & 0.66 & 0.19 \\
      \midrule
      \multicolumn{7}{l}{\textit{Unified tokenizers}} \\
      UniTok~\citep{ma2025unitok} & 256 & 256 & 0.38 & - & - & - \\
      TokenFlow~\citep{qu2025tokenflow} & 384 & 729 & 0.63 & 22.77 & 0.73 & - \\
      \model{} & 512 & 256 & 0.54 & 30.77 & 0.62 & 0.14 \\
      \model{} $\dagger$ & 512 & 256 & 0.38 & 31.09 & 0.64 & 0.12 \\
      \bottomrule
      \multicolumn{7}{l}{$\dagger$ denotes using semantic decoder after joint pre-training.}
    \end{tabular}
    \label{tab:image-recon}
    
  \end{minipage}%
  \hfill
  \begin{minipage}[t]{0.45\textwidth}
    \scriptsize
    \captionsetup{justification=raggedright, singlelinecheck=false}
    \caption{\textbf{Image editing performance on GEdit-Bench (EN)~\citep{liu2025step1x-edit}.} Metrics are evaluated by GPT-4.1.}
    \centering
    \setlength{\tabcolsep}{7pt}
      \begin{tabular}{@{}lccc@{}}
        \toprule
        \textbf{Model} & \textbf{G\_SC} & \textbf{G\_PQ} & \textbf{G\_O} \\ 
        \midrule
        \multicolumn{4}{c}{\textit{Specialized image editing model}} \\
        Instruct-P2P~\citep{brooks2023instructpix2pix} & 3.58 & 5.49 & 3.68 \\
        MagicBrush~\citep{zhang2023magicbrush} & 4.68 & 5.66 & 4.52 \\
        AnyEdit~\citep{yu2025anyedit} & 3.18 & 5.82 & 3.21 \\
        UniWorld-V1~\citep{lin2025uniworld} & 4.93 & 7.43 & 4.85 \\
        OmniGen~\citep{xiao2025omnigen} & 5.96 & 5.89 & 5.06 \\
        OmniGen2~\citep{wu2025omnigen2} & 7.16 & 6.77 & 6.41 \\
        Step1X-Edit~\citep{liu2025step1x-edit} & 7.09 & 6.76 & 6.70 \\ 
        \midrule
        \multicolumn{4}{c}{\textit{Unified model}} \\
        \rowcolor{gray}
        BAGEL*~\citep{deng2025bagel} & 7.36 & 6.83 & 6.52 \\
        Ours (single-round) & 6.04 & 6.86 & 5.54 \\
        Ours (multi-round) & 6.60 & 6.25 & 5.78 \\ 
        \bottomrule
        \multicolumn{4}{l}{* indicates model pretrained on large-scale interleaved data.}
      \end{tabular}
    \label{tab:gedit}
  \end{minipage}
\end{table*}

To evaluate the image editing performance of our model, we use GEdit-Bench-EN~\citep{liu2025step1x-edit}, a benchmark featuring real-world user instructions across 11 diverse categories. Performance is measured using three metrics: Semantic Consistency (SC), Perceptual Quality (PQ), and Overall Score (O), all on a 0–10 scale.
Since our model does not rely on large-scale interleaved pre-training, we find consistent resolution between understanding and generation stages to be critical for effective editing. 
Therefore, we report results using a base model trained without mixed-resolution strategy.
As shown in Table~\ref{tab:gedit}, our model achieves competitive single-round editing quality (G\_PQ) compared to existing methods, while demonstrating strong multi-round success rates (G\_SC). Although the overall score lags behind prior work, we attribute this gap primarily to two factors: the absence of large-scale multimodal sequence pre-training, and the high per-token detail density in our continuous tokenizer—both of which limit current fidelity under complex instructions. We discuss these limitations in depth in Sec.~\ref{sec:discussion} and plan to address them in future work.

\subsection{Image Reconstruction}

We compare the reconstruction performance of our model with existing tokenizers in Table~\ref{tab:image-recon}. \model{} operates at a 32$\times$ compression ratio, encoding 512$\times$512 images into a compact representation of 256 continuous latent tokens. Under this high compressionm \model{} achieves an rFID of 0.54 and a PSNR of 30.77 db, indicating strong structural alignment and high pixel fidelity. After the semantic decoder is jointly trained during the pre-training process of the unified multi-modal model, the reconstruction quality is further improved, with LPIPS decreasing to 0.12 and rFID dropping to 0.38. This suggests that end-to-end optimization within the unified framework enhances the semantic decoder's ability to retain fine textures and global semantics. 

\section{Analysis}
In this section, we analyze some critical designs in our \unifiedmodel{}. Here, we switch to smaller data scale during pre-training compared to previous experiments, and a smaller dense language model, \textit{i.e.,} Qwen-2.5-3B~\citep{qwen25}, for efficient ablation analysis.

\subsection{Unified Representation for Competing Tasks}

In this section, we explore the influence of unified visual representation on the understanding and generation capability of unified multi-modal models.
For a comprehensive comparison, three tokenizers are introduced into our experiments: CLIP for image understanding, VAE for image generation, and the proposed \model{}, which can be used either for understanding or generation. By combining these tokenizers, we get four kinds of unified multi-modal models for image understanding and generation, to be specific: 1) CLIP as und\_tok (short for understanding tokenizer) and VAE as gen\_tok (short for generation tokenizer), 2) CLIP as und\_tok and \model{} as gen\_tok, 3) \model{} as und\_tok and VAE as gen\_tok, and 4) \model{} as both und\_tok and gen\_tok. It's noteworthy that the former three have separate understanding and generation representation spaces while the last one has unified representation space.

\textbf{Unified representation matters for understanding.} As can be seen in Table.~\ref{tab:mingtok_beats_clip_and_vae}, the last row \model{} as both und\_tok and gen\_tok performs the best on `Average', which implies that pretraining in a unified representation space has better image understanding performance than that in two separate spaces. To look further, within each und\_tok group, the performance is inferior when VAE is used as gen\_tok. 
The reason is speculated to be that during joint training, the MLLM has to expend considerable effort to align the understanding and generation representation spaces. 
From this perspective, since VAE's features focus more on details and have little semantic information, while \model{}'s features themselves contain sufficient semantic information, so it is easier for \model{} than VAE to align with und\_tok's understanding representations.

\begin{table}[t]
\small
\centering
\caption{
Ablation study on visual representation designs for unified multi-modal models. Und\_Tok and Gen\_Tok denote the tokenizers used for understanding and generation, respectively. When \model{} serves as both the understanding and generation tokenizer, the model achieves optimal performance on both understanding and generation tasks, demonstrating the advantage of a unified, shared visual space.
}
\begin{tabular}{ll|ccccccc|c}
\toprule
Und\_Tok & Gen\_Tok & MMB & MMStar & MMMU & Mathvista & AI2D & OCRBench & Average $\uparrow$ & GenEval $\uparrow$ \\
\midrule
CLIP & VAE & 66.41 & 51.03 & 40.00 & 48.0 & 69.98 & 36.1 & 51.92  &  0.3591 \\
CLIP & \model{} &	67.18 & 52.13 &	38.67 &	49.5 &	72.25 &	35.5 &	52.54 & 0.4600 \\
\model{} & VAE &	69.42 &	51.57 & 40.89 &	47.9 &	71.99 &	30.4 &	52.03 & 0.3950 \\
\model{} & \model{} &	69.93 &	51.89 &	40.44 &	52.3 & 	75.23 &	31.6 &	\textbf{53.57} & \textbf{0.4654}  \\
\bottomrule
\end{tabular}
\label{tab:mingtok_beats_clip_and_vae}
\end{table} 
\begin{figure}[t]
\centering
\includegraphics[width=\textwidth]{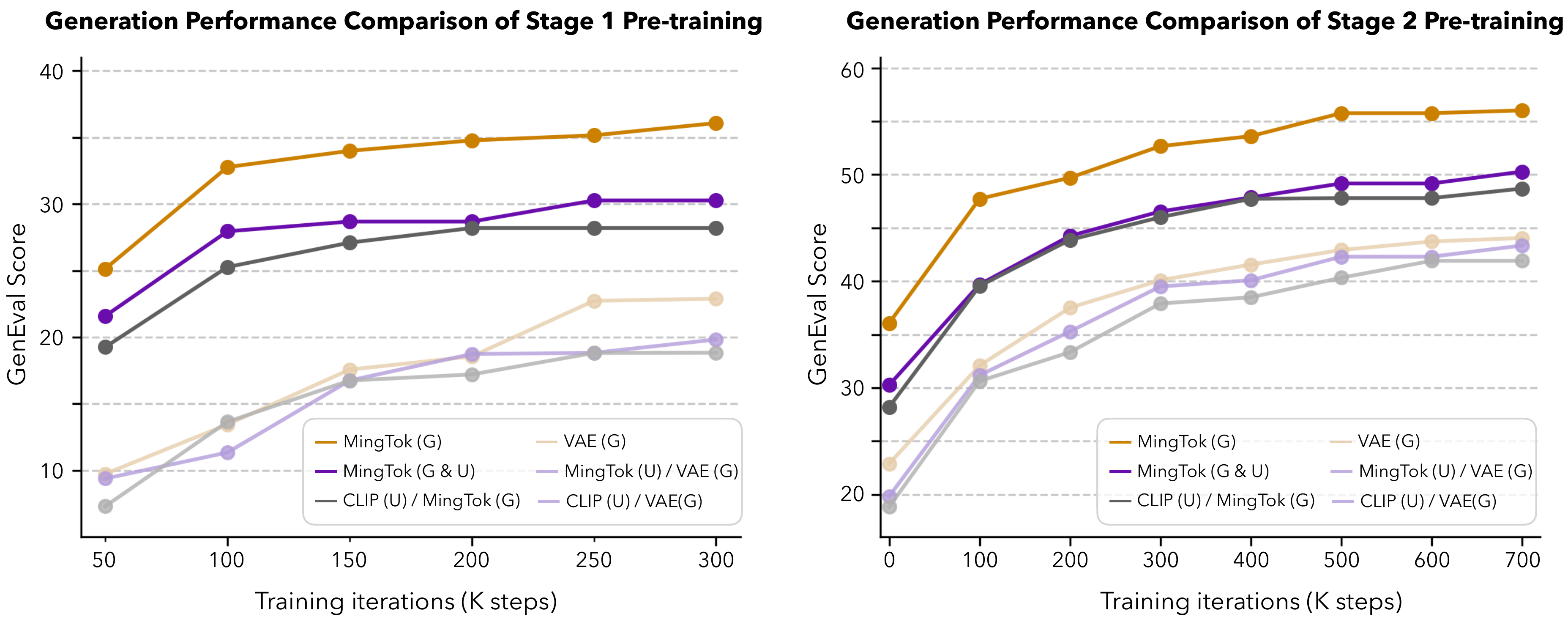}
\caption{
Generation performance comparison during pre-training across different understanding (U) and generation (G) tokenizer combinations. Using \model{} as the generation representation (\model{} (G)) achieves the best performance in generation-only training, significantly outperforming VAE-based representations (VAE (G)). When \model{} is used for both roles (\model{} (G \& U), unified setting), the performance gap between pure generation and unified training narrows notably, demonstrating the benefit of universal visual representations for joint vision-language modeling.
}
\label{fig:generation_s1s2_curves}
\end{figure}

\textbf{Unified representation matters for generation.} First of all, \model{} demonstrates its ability to image generation task in Table.~\ref{tab:mingtok_beats_clip_and_vae}, as we can see that no matter what und\_tok is, \model{} as gen\_tok always shows a significant `GenEval' improvement compared to its VAE counterpart. We hypothesize that this advantage may come from the fact that \model{}' features, not only contain the detail information used for image reconstruction but also contain sufficient semantic information, which can probably accelerate the convergence of image generation. So in this way, \model{} can serve as an alternative to existing image generation tokenizers. Besides, the best generation performance is also obtained when \model{} acts as both und\_tok and gen\_tok, which indicates that pretraining in a unified
representation space is more effective for image generation tasks than that in different representation spaces. 

To further explore the effect of \model{} as an image generation tokenizer, 
we study the training process of the generation tasks, as illustrated in Figure.~\ref{fig:generation_s1s2_curves}, with two additional settings: pure generation with \model{} (denoted as MingTok(G)), and pure generation with VAE (denoted as VAE(G)). From these curves, it's easy to draw the following conclusions: 
1) Generation-only models achieve superior performance than joint trained models with both understanding and generation capabilities;
2) \model{} outperforms VAE as an image generation tokenizer; 3) Joint training in a unified representation space minimizes performance degradation in image generation tasks.

\subsection{Multi‑round Understanding, Generation, and Editing}
\label{sec:multiround-eval}

As discussed in Sec.~\ref{sec:multi-round}, most existing unified architectures either do not support explicit multi‑round training, or—when extended to such scenarios—must maintain tokens from multiple heterogeneous feature spaces (separately optimised for understanding and for generation) in memory simultaneously. This heterogeneity not only complicates the attention mechanism during multi‑round training, but also makes the sequential editing process more cumbersome to optimise.  

To examine how task formulation impacts multi‑round performance, we begin with two foundational comparison experiments:

\begin{itemize}
    \item \textbf{Recon + Edit (Baseline):} A standard single‑round setting in which the model reconstructs the original image and then performs a single edit.
    \item \textbf{Add Seg‑as‑Edit (Proposed):} Extends the baseline by adding a \emph{reconstruction + segmentation} editing task. Specifically, a portion of the training samples are modified to require reconstruction followed by a segmentation‑as‑editing operation, encouraging the model to learn fine‑grained boundary localisation and semantic consistency within its latent space.
\end{itemize}

\begin{table}[tp]
\centering
\captionsetup{justification=centering}
\caption{Multi‑round editing performance: Baseline vs.\ Seg‑as‑Edit.}
\label{tab:multi-round-editing}
\small
\begin{tabular}{@{}l|ccc|ccc@{}}
\toprule
\multirow{2}{*}{Category} & \multicolumn{3}{c|}{Recon+Edit} & \multicolumn{3}{c}{Add Seg‑as‑Edit} \\
\cmidrule(lr){2-4} \cmidrule(lr){5-7}
 & G\_SC & G\_PQ & G\_O & G\_SC & G\_PQ & G\_O \\
\midrule
background\_change & 6.931 & 6.345 & 6.177 & 7.448 & 6.138 & 6.563 \\
color\_alter       & 8.118 & 6.588 & 6.069 & 8.118 & 5.912 & 6.595 \\
material\_alter    & 5.536 & 5.714 & 4.989 & 5.214 & 5.929 & 4.651 \\
motion\_change     & 2.818 & 6.818 & 2.741 & 3.636 & 6.727 & 3.459 \\
ps\_human          & 3.756 & 8.268 & 4.047 & 4.171 & 8.146 & 4.429 \\
style\_change      & 7.167 & 5.042 & 5.788 & 7.271 & 5.292 & 6.032 \\
subject‐add        & 6.026 & 7.684 & 5.674 & 6.605 & 7.289 & 6.208 \\
subject‐remove     & 8.690 & 7.262 & 7.542 & 8.857 & 7.238 & 7.747 \\
subject‐replace    & 6.891 & 5.913 & 5.987 & 7.609 & 5.957 & 6.418 \\
text\_change       & 3.938 & 5.210 & 3.637 & 4.469 & 5.185 & 4.238 \\
tone\_transfer     & 7.800 & 7.040 & 7.005 & 7.520 & 6.760 & 6.894 \\
\midrule
\textbf{Average}   & \textbf{6.034} & \textbf{6.535} & \textbf{5.423} & \textbf{6.447} & \textbf{6.416} & \textbf{5.749} \\
\bottomrule
\end{tabular}
\end{table}

\begin{figure*}[t] 
\centering
\includegraphics[width=0.95\textwidth]{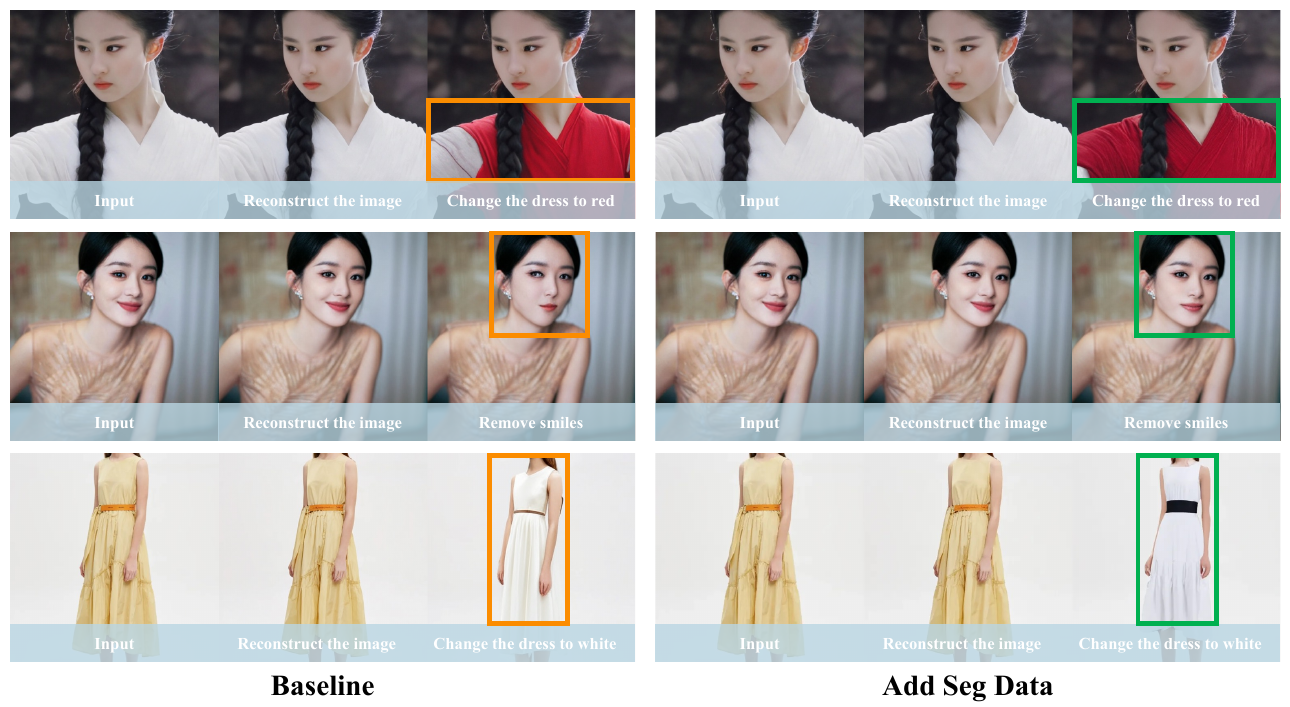}
\caption{
Qualitative comparison of multi-step editing across the three training strategies. 
\textbf{Column~1 (Baseline):} Struggles with sequential edits, altering the shirt's style during an incomplete color change and leaving artifacts after smile removal.
\textbf{Column~2 (Add Seg-as-Edit):} Benefiting from the segmentation task, our main model improves color fidelity and preserves the shirt's style; smile removal is cleaner.
}
\label{fig:edit_examples}
\end{figure*}

\paragraph{Analysis.}  
Across categories, Seg‑as‑Edit improves semantic consistency in 9/11 tasks, with the largest gains in motion\_change (+0.82 G\_SC) and background\_change (+0.52 G\_SC). The average \texttt{G\_SC} rises +0.41, and \texttt{G\_O} rises +0.33, indicating better preservation of target semantics and overall output quality. Perceptual quality remains on par in most categories, reflecting that structural regularisation via segmentation strengthens consistency without sacrificing visual fidelity. Qualitative examples in Fig.~\ref{fig:edit_examples} further show that styles and local details are better maintained over successive edits.

Such targeted improvements in multi‑round robustness form a natural bridge to practical, real‑world workflows, where editing often proceeds through long, dependent sequences of transformations.

\paragraph{Qualitative Analysis of Practical Workflows.}
Beyond controlled ablations, the strengths of our unified architecture are most evident in complex sequential scenarios typical of creative practice — the very situations where prior art suffers from feature‑space fragmentation and attention complexity. In popular single‑round editors such as \textbf{Qwen‑Image}, each edit is treated independently (\emph{stateless} editing), making it difficult to maintain \emph{identity consistency}: facial features or clothing can drift noticeably between iterations. In contrast, our method preserves a coherent latent representation across steps, ensuring consistency even in extended editing chains.

\begin{figure}[t]
\centering
\includegraphics[width=0.95\textwidth]{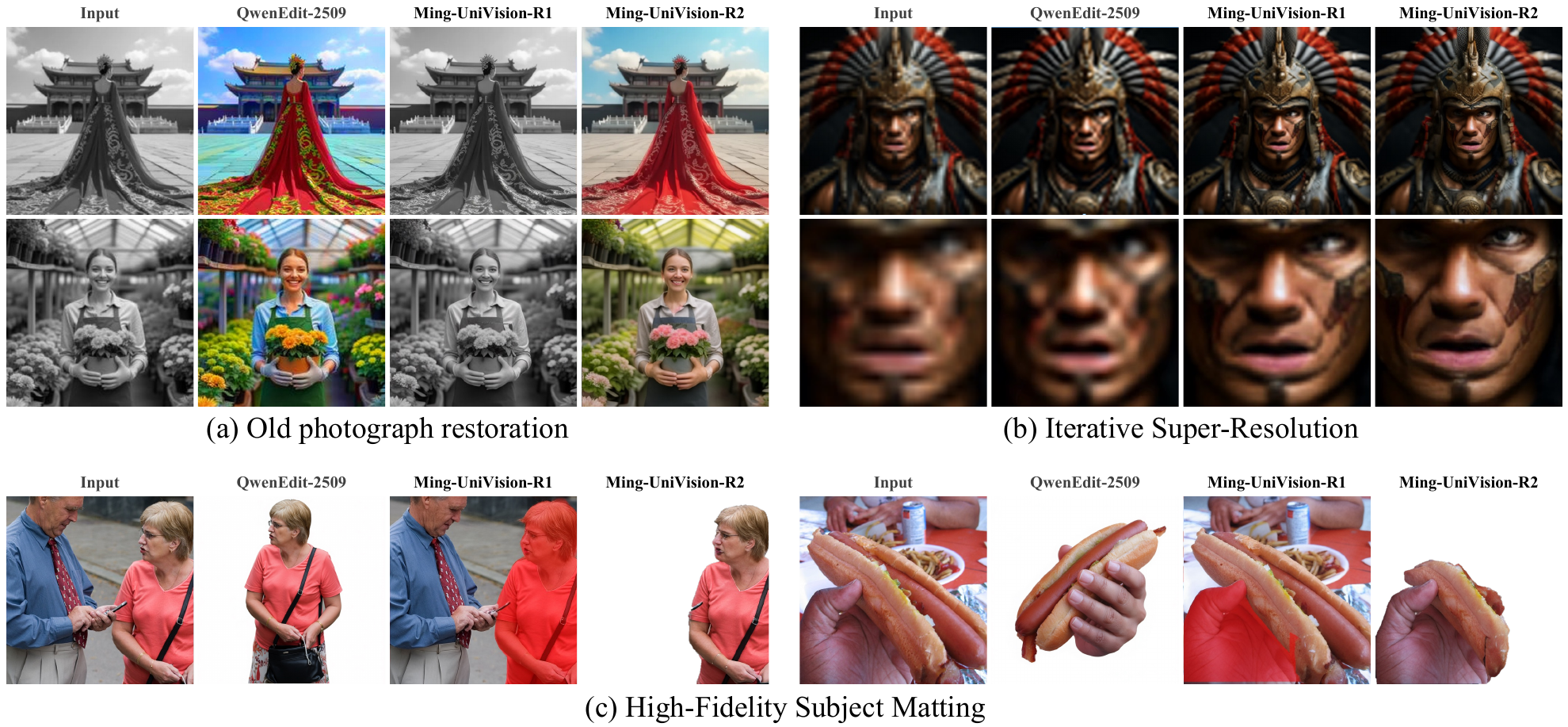}
\caption{Qualitative illustration of multi-step visual editing workflows enabled by our unified model. (a) An old photograph is restored through a sequential process: first upscaled to higher resolution, then colorized. (b) Further iterative super-resolution can be applied to progressively enhance image quality, demonstrating the model’s capability for context-preserving, in-context refinement. (c) High-fidelity subject matting is achieved in two steps: segmentation to highlight the subject, then background removal.}
\label{fig:qual-photo-restoration-super-res}
\end{figure}

\begin{itemize}
    \item \textbf{Old Photograph Restoration:} Multi‑stage processes like super‑resolution followed by colourisation often accumulate errors. Our model transitions seamlessly from resolution enhancement to plausible colouring within a stable latent space (Fig.~\ref{fig:qual-photo-restoration-super-res}), preserving fine detail and scene coherence.
    \item \textbf{Iterative Super‑Resolution:} In traditional methods, repeated super‑resolution magnifies artifacts. \model{} maintains structure and introduces plausible details across iterations (\emph{low‑res} $\rightarrow$ \emph{super‑res} $\rightarrow$ \emph{super‑res}), avoiding quality degradation cascades.
    \item \textbf{High‑Fidelity Subject Matting:} Leveraging segmentation‑as‑editing to create a high‑quality alpha mask prior to background removal minimises halo artifacts and retains delicate structures like individual hair strands, as in Fig.~\ref{fig:qual-photo-restoration-super-res}.
\end{itemize}

These practical cases highlight the core benefit of \model{}’s stateful design: by unifying understanding and generation in a single feature space, it converts fragmented, multi‑step edits into a coherent creative flow, enabling more powerful and reliable interactive tools.


\subsection{Visualized CoT for Visual Reasoning and Image Editing}

Building on the multi-image generation capabilities of \model{} described in Sec.~\ref{sec:multi-round}, we introduce Visualized Chain-of-Thought (Visualized CoT), 
which enables explicit visual reasoning by generating intermediate visualizations prior to image editing.
This novel paradigm first highlights the regions of the reference image that require modification according to the editing instructions.
Subsequently, the edited image is generated guided by these visual cues.

Unlike Generation Chain-of-Thought (GoT) \citep{fang2025got}, which conducts reasoning in natural language, 
Visualized CoT performs reasoning visually by directly generating an image in which the regions to be edited are highlighted with overlayed colors.
Moreover, GoT requires translating its reasoning outputs into an editing mask, 
which is then encoded to condition the image editing process. 
In contrast, our method directly leverages the visualized context to guide the editing.
This end-to-end visual reasoning and generation framework enables seamless integration of understanding and editing, enhancing both transparency and efficiency in the image editing workflow.

\begin{figure}[ht]
\centering
\includegraphics[width=0.95\textwidth]{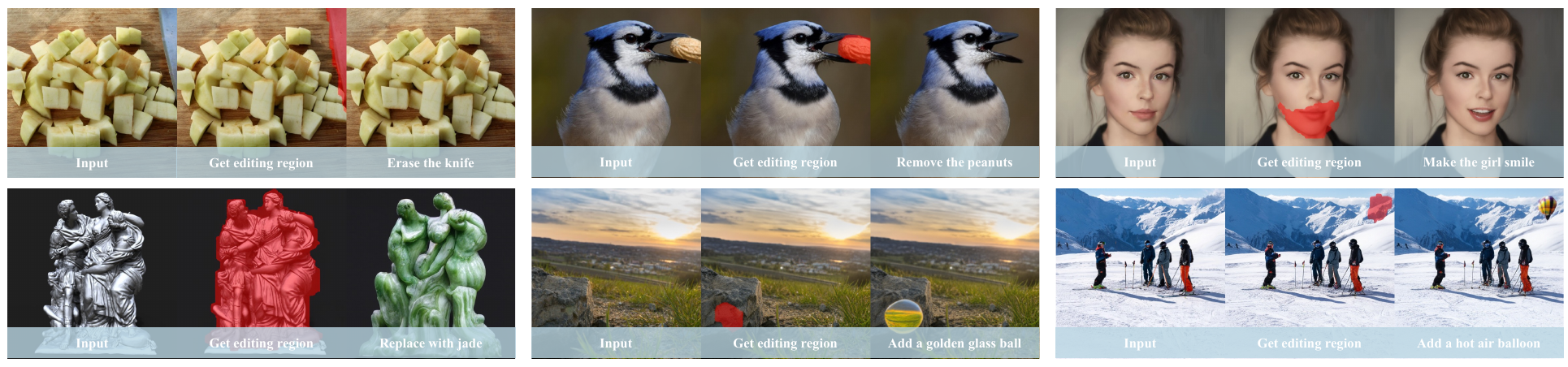}
\caption{Illustration of the Visualized CoT. Under this paradigm, the model first performs visual reasoning to generate an image in which the regions requiring editing are highlighted on the reference image. Subsequently, the model completes the image editing according to these visual cues.}
\label{fig:got_process}
\end{figure}

\begin{table}[t]
\centering
\captionsetup{justification=centering}
\caption{Quantitative comparison between single-step editing and two-step generative Visualized CoT approach.}
\small
\label{tab:multi_round}
\begin{tabular}{
@{}l|p{1.1cm}<{\centering} p{1.1cm}<{\centering} p{1.1cm}<{\centering}|p{1.1cm}<{\centering} p{1.1cm}<{\centering} p{1.1cm}<{\centering}@{}}
\toprule
\multirow{2}{*}{Method} & \multicolumn{3}{c|}{GEdit-Bench-EN-Sub} & \multicolumn{3}{c}{GEdit-Bench-EN-Full} \\
                           & G\_SC       & G\_PQ       & G\_O        & G\_SC       & G\_PQ       & G\_O        \\ \midrule
Single-step                & 6.055       & 6.890       & 5.582       & 6.042       & 6.855       & 5.535       \\
Visualized CoT             & 6.543       & 6.278       & 5.791       & 6.490       & 6.259       & 5.668       \\ \bottomrule
\end{tabular}
\end{table}

\textbf{To construct the training data}, we follow UniWorld-V1~\citep{lin2025uniworld} to obtain the editing regions by calculating the differences between edited images and their reference images. 
The resulting editing region masks are then overlaid onto the reference images as intermediate outputs for visual reasoning. 
The edited images are then used as the final desired outputs, forming a two-step Visualized CoT image editing paradigm. 
The unified feature space for visual generation and understanding enables this multi-image generation framework to be trained end-to-end.

\textbf{To evaluate the effectiveness of this new paradigm}, we conduct quantitative evaluations on the GEdit-Bench~\citep{liu2025step1x-edit}. 
We compare our method against a standard single-step pipeline, which serves as a baseline by directly generating the edited image from the reference image and textual instructions.
As show in Table~\ref{tab:multi_round}, 
our Visualized CoT approach outperforms the single-step baseline, most notably in semantic consistency, with a gain of +0.5 points.
This improvement is attributed to the intermediate visual reasoning result, which introduces a strong spatial prior and reduces editing ambiguity. 
Figure~\ref{fig:got_process} illustrates the visual reasoning process of Visualized CoT, where the model accurately identifies the regions requiring editing.

\subsection{Discussions}
\label{sec:discussion}

This work represents an early step toward unified vision-language modeling in continuous latent spaces.
Our goal in releasing this version is not to claim state-of-the-art performance across all tasks, but to highlight the potential of using a single, shared continuous representation for both visual understanding and autoregressive generation. 
While \model{} and \unifiedmodel{} demonstrate promising capabilities in joint perception and synthesis, it still has its limitations, particularly in fine-grained editing and understanding. We plan to address these in subsequent versions.

\textbf{Limitations on current editing performance. } While our model supports flexible in-context image editing through autoregressive sequence modeling, its performance on quantitative editing benchmarks has room for improvement. We identify two key factors that currently limit editing success rate and quality. 

First, the model lacks large-scale \textit{interleaved-pretraining}, \textit{i.e.,} pre-training on sequences that alternate between text and image tokens across diverse editing scenarios. Such data could help the model learn generalizable edit patterns before fine-tuning. Without it, the model relies heavily on SFT to acquire editing behavior, which may not generalize well beyond seen prompts. This limitation is particularly pronounced under mixed-resolution training, where generation and editing operate at lower resolutions and thus cannot leverage the understanding capability learned during high-resolution understanding training. Moving forward, achieving optimal joint performance in a unified resolution setting will be a key focus of our next-stage development.

Second, due to the high compression ratio of \model{} designed for generation efficiency, each latent token encodes a large amount of visual detail. This high information density makes fine-grained editing challenging, as small changes in tokens can lead to significant and often uncontrollable changes in pixel space. In future work, we plan to explore higher-resolution tokenizeation or lower compression ratios to reduce per-token information load, thereby imprving the precision and quality in both generation and editing.

\textbf{Challenges in multi-round and freeform interleaved interaction. }
While \model{} supports basic in-context editing, it still falls short in more advanced interaction patterns. In multi-round editing, we observe that the model struggles to generalize to editing sequences longer than those seen during training. More fundamentally, the model yet still struggles with freeform interleaved understanding and generation (or editing), such as arbitrarily ordered sequences like "describe, generate, compare, revise, regenerate, \textit{etc.}". The current training paradigm focused on structured, unidirectional flows, does not sufficiently prepare the model for flexible, dynamic task switching. We leave the exploration of length-generalizable editing strategies and training on diverse, real-world interaction trajectories for future work. 

\textbf{Mutual enhancement between generation and understanding. }
Finally, we emphasize that a unified visual representation is not merely an architectural choice, but a key enabler for mutual enhancement between generation and understanding. 
By sharing the same representation space across tasks, \model{} allows knowledge learned in generation—such as fine-grained texture synthesis and compositional reasoning—to benefit perception, while visual understanding provides grounded, coherent priors for more controllable and faithful generation.
We observe early evidence of this synergy: using a shared representation reduces the performance gap between pure generation and unified training, and alleviates task competitions that typically arise when pathways diverge.
We hope this perspective inspires the research community to further explore unified modeling of generation and understanding, moving toward more integrated and synergistic multimodal systems.
\section{Contributors}
\label{sec:contri}

Ziyuan Huang, DanDan Zheng, Cheng Zou, Rui Liu, Xiaolong Wang, Kaixiang Ji, Weilong Chai, Jianxin Sun, Libin Wang, Yongjie Lv, Taozhi Huang, Jiajia Liu, Qingpei Guo, Ming Yang, Jingdong Chen, Jun Zhou

\clearpage

\newpage

\bibliographystyle{assets/plainnat}
\bibliography{ref}

\end{document}